\documentclass[11pt,a4paper]{article}
\usepackage[hyperindex,breaklinks]{hyperref}
\usepackage[hyperref]{acl2019}
\usepackage{times}
\usepackage{latexsym}
\usepackage{enumitem}
\usepackage{booktabs}
\usepackage{tikz}
\usepackage{amsmath}
\usepackage{tabularx}
\usepackage{filecontents}
\usepackage{multirow}
\usepackage{multicol}
\usepackage{hhline}
\usepackage[skip=2pt]{caption}
\usepackage{color}
\usepackage{colortbl}
\usepackage{pgfplots}

\usetikzlibrary{
  shadings,%
  shadows,%
  trees,%
  shapes,%
  arrows,%
  positioning,%
  calc,%
  automata,%
  matrix,%
  arrows.meta%
}
\usepackage{tikz-dependency}

\usepackage{array}
\newcolumntype{L}[1]{>{\raggedright\let\newline\\\arraybackslash\hspace{0pt}}m{#1}}
\newcolumntype{C}[1]{>{\centering\let\newline\\\arraybackslash\hspace{0pt}}m{#1}}
\newcolumntype{R}[1]{>{\raggedleft\let\newline\\\arraybackslash\hspace{0pt}}m{#1}}

\definecolor{cb-black}      {RGB}{  0,   0,   0}
\definecolor{cb-blue-green} {RGB}{  0,  073,  073}
\definecolor{cb-green-sea}  {RGB}{  0, 146, 146}
\definecolor{cb-rose}       {RGB}{255, 109, 182}
\definecolor{cb-salmon-pink}{RGB}{255, 182, 119}
\definecolor{cb-purple}     {RGB}{ 73,   0, 146}
\definecolor{cb-blue}       {RGB}{ 0, 109, 219}
\definecolor{cb-lilac}      {RGB}{182, 109, 255}
\definecolor{cb-blue-sky}   {RGB}{109, 182, 255}
\definecolor{cb-blue-light} {RGB}{182, 219, 255}
\definecolor{cb-burgundy}   {RGB}{146,   0,   0}
\definecolor{cb-brown}      {RGB}{146,  73,   0}
\definecolor{cb-clay}       {RGB}{219, 209,   0}
\definecolor{cb-green-lime} {RGB}{ 36, 255,  36}
\definecolor{cb-yellow}     {RGB}{255, 255, 109}

\usepackage{url}

\aclfinalcopy % Uncomment this line for the final submission
\def\aclpaperid{31} %  Enter the acl Paper ID here

\newcommand{\specialcell}[2][c]{\begin{tabular}[#1]{@{}c@{}}#2\end{tabular}}

\title{A Systematic Comparison of English Noun Compound Representations}

\author{Vered Shwartz\\
  Computer Science Department, Bar-Ilan University, Ramat-Gan, Israel\\
  {\tt vered1986@gmail.com}}

\date{}

\begin{document}
\maketitle
\begin{abstract}
     Building meaningful representations of noun compounds is not trivial since many of them scarcely appear in the corpus. To that end, composition functions approximate the distributional representation of a noun compound by combining its constituent distributional vectors. In the more general case, phrase embeddings have been trained by minimizing the distance between the vectors representing paraphrases. We compare various types of noun compound representations, including distributional, compositional, and paraphrase-based representations, through a series of tasks and analyses, and with an extensive number of underlying word embeddings. We find that indeed, in most cases, composition functions produce higher quality representations than distributional ones, and they improve with computational power. No single function performs best in all scenarios, suggesting that a joint training objective may produce improved representations. 
\end{abstract}

\section{Introduction}
\label{sec:intro}
The simplest way to obtain a vector representation for a multiword term is to treat it as a single token, e.g. by replacing spaces with underscores, and train a standard word embedding algorithm. This is typically done for common n-grams, which often include named entities (e.g. New York), but in theory can also be based on syntactic criteria, for instance in order to learn noun compound vectors. The main issue with this approach is that word embedding algorithms require sufficient term frequency to obtain meaningful representations, and many noun compounds rarely occur in text corpora \cite{kim2006interpreting}. 

To overcome the sparsity issue, it is common to learn a composition function which computes a noun compound vector from its constituents' distributional representations, e.g. vec(\textit{cost estimate}) = f(vec(\textit{cost}), vec(\textit{estimate})). Various functions have been proposed in the literature, typically based on vector arithmetics \cite[e.g.][]{mitchell2010composition,zanzotto2010estimating,dinu-pham-baroni:2013:CVSC}. Such functions are learned with the objective of minimizing the distance between the observed (distributional) vector and the composed vector of each noun compound, and most functions are limited to binary noun compounds. 

A parallel line of work computes phrase embeddings for variable-length phrases, by adapting the word embedding training objective \cite{poliak-EtAl:2017:EACLshort} or by minimizing the distance between the representations of paraphrases \cite{wieting2015towards,wieting-17-millions,wieting-mallinson-gimpel:2017:EMNLP2017}. Paraphrase-based phrase embeddings require a large number of paraphrases as training instances. Such paraphrases are often generated by translating an English phrase into a foreign language and back to English, considering variations in translation as paraphrases. This technique is referred to as ``bilingual pivoting'' or ``backtranslation'' \cite{P01-1008,bannard-callison-burch-2005-paraphrasing,N13-1092,mallinson-sennrich-lapata:2017:EACLlong}. 

In this work we test the quality of noun compound representations produced by different methods, including distributional representations, composition functions, and paraphrase-based phrase embeddings. We extend the work of \newcite{W16-1604}, who evaluated various composition functions on the noun compound relation classification task, in several aspects. First, we test a broader range of representations, which may differ both in their architectures and in their training objectives. Second, we train each representation with a wide variety of underlying word embeddings, and analyze the representation's behaviour across the different word embeddings. Finally, we use several tasks to evaluate the representation quality: relation classification (what is the relationship between the constituents?),  property classification (is a \textit{cheese wheel} round?), as well as a qualitative and quantitative analysis of the nearest neighbours. The results confirm that the distributional representations of rare noun compounds are indeed of low quality. Across representations, the nearest neighbours of a target noun compound vector typically include many trivial similarities such as other noun compounds with a shared constituent. 

Among the composition functions, functions with more computational power and parameters generally produced higher quality representations. The paraphrase-based functions outperformed the others in the property prediction task, while the compositional functions performed better on relation classification. The results suggest that learning a composition function with a combined training objective is a promising research direction that may result in improved noun compound representations.\footnote{The code and data is available at \url{https://github.com/vered1986/NC_Embeddings}.}

\section{Representations}
\label{sec:representations}
We trained 315 distributional semantic models (DSMs) that differ by their training objective (Section~\ref{sec:rep_training_objective}) and the underlying embeddings used for the constituent nouns (Section~\ref{sec:rep_constituent_embeddings}). 

\subsection{Training Objective}
\label{sec:rep_training_objective}

\paragraph{Distributional.} This approach simply treats a noun compound as a single token w$_1$\_w$_2$, and learns standard word embeddings for the words and noun compounds in the corpus. 

\paragraph{Compositional.} We learn a function $f(\cdot, \cdot) : \mathcal{R}^d \times \mathcal{R}^d \rightarrow \mathcal{R}^d$ which, for a given noun compound, operates on the word embeddings of its constituent nouns, and returns a vector representing the compound. Following \newcite{W16-1604} and earlier work, the training objective is to minimize the distance between the observed distributional embedding $\vec{v}_{w_1\_w_2}$ and the composed vector $f(\vec{v}_{w_1}, \vec{v}_{w_2})$. 

We train the following composition functions: 

\begin{itemize}[leftmargin=*]
 \item \textbf{Add} \cite{mitchell2010composition}: $f(\vec{v}_{w_1}, \vec{v}_{w_2}) =$ $\alpha \vec{v}_{w_1} + \beta \vec{v}_{w_2}$, $\alpha, \beta$ are scalars. 
 \item \textbf{FullAdd} \cite{zanzotto2010estimating,dinu-pham-baroni:2013:CVSC}: $f(\vec{v}_{w_1}, \vec{v}_{w_2}) =$ $W_1 \vec{v}_{w_1} + W_2 \vec{v}_{w_2}$, where $W_1, W_2 \in \mathcal{R}^{d \times d}$ are matrices.  
 \item \textbf{Matrix} \cite{W16-1604}: $f(\vec{v}_{w_1}, \vec{v}_{w_2}) = tanh(W \cdot [\vec{v}_{w_1}; \vec{v}_{w_2}])$, where $W \in \mathcal{R}^{2d \times d}$. This is the application of the recursive matrix-vector method of \newcite{D12-1110} to binary phrases.\footnote{Originally, this method was trained with an extrinsic training objective of sentiment analysis.}
 \item \textbf{LSTM}: encoding the compound with a long short-term memory network \cite[LSTM; ][]{hochreiter1997long}: $f(\vec{v}_{w_1}, \vec{v}_{w_2}) = LSTM(\vec{v}_{w_1}, \vec{v}_{w_2})$.
 \end{itemize}

\paragraph{Paraphrase-based.} In this approach we follow the literature of paraphrase-based phrase embeddings \cite[e.g.][]{wieting2015towards,wieting-mallinson-gimpel:2017:EMNLP2017}. We generate paraphrases for each noun compound, and train the function with the objective of producing similar vectors to the noun compound and its paraphrase. 

To obtain the representation of a phrase (either a noun compound or its variable-length paraphrase), we encode it with an LSTM. For a given noun compound NC = w$_1$ w$_2$ and its paraphrase $p$, we set the loss to: 

\begin{equation*}
\vspace*{-5pt}
max(0, \lambda - cos(v_{NC}, v_{p}) + cos(v_{NC}, v_{p'}))    
\end{equation*}

\noindent where $v_{x} = LSTM(x)$ is the encoding of phrase x, p' is a negative-sampled paraphrase, and $\lambda$ was set to 0.6 based on its value in \newcite{wieting2015towards}. The following approaches were used to obtain the paraphrases:

\begin{itemize}[leftmargin=*]
 \item \textbf{Backtranslation}: We translate each noun compound to foreign language(s) and back to English, as in \newcite{wieting-mallinson-gimpel:2017:EMNLP2017}. Specifically, we use the DeepL Translator web interface,\footnote{\url{https://www.deepl.com}} performing translation from English to 4 different foreign languages (French, Italian, Spanish, and Romanian) and back to English. We focused on Romance languages because they translate English noun compounds to noun phrases with prepositions \cite{girju:2007:ACLMain}, and we were hoping that this would drive the backtranslation to be more explicit. For example, 
\textit{baby oil} is translated in French to \textit{huile pour b\'eb\'e}, which literally means \textit{oil for baby}. In practice, translating back to English mostly generates paraphrases which are other noun compounds (synonyms or related terms), rather than prepositional paraphrases. 

We use all the suggested translations to generate a large list of paraphrases for each noun compound, but we apply two filters. First, we trivially remove the noun compound itself from its list of paraphrases. Second, the translation sometimes yields non-English phrases (a result of an error in the translation), which we automatically identify and remove using a language identification tool.\footnote{\url{https://pypi.org/project/guess_language-spirit/}} After filtering around half of the paraphrases, we remain with an average number of 6.71 paraphrases per compound. 

 \item \textbf{Co-occurrence}: We treat the frequent joint occurrences of w$_1$ and w$_2$ in a corpus as paraphrases, e.g. \textit{apple cake} may yield a paraphrase like ``\textit{cake made of apples}''. Specifically, we use the paraphrases obtained by \newcite{shwartz-dagan-2018-paraphrase} from the Google N-gram corpus \cite{brants2006web}. The paraphrases are of variable length (3-5 words), and have been pre-processed to remove punctuation, adjectives, adverbs and determiners. The averaged number of paraphrases per compound is 9.18. 

\end{itemize}

\begin{table*}[!t]
    \centering
    \small
    \begin{tabular}{cccc}
\toprule
\multicolumn{4}{c}{\textit{syndicate representative} (rare)} \\ \midrule
\rowcolor{lightgray} \multicolumn{4}{c}{\textbf{Distributional}} \\
\multicolumn{4}{c}{geloios} \\
\multicolumn{4}{c}{t.franse} \\
\multicolumn{4}{c}{adopter(s} \\
\multicolumn{4}{c}{ahchie} \\
\multicolumn{4}{c}{anquish} \\
\midrule
\rowcolor{lightgray} \multicolumn{4}{c}{\textbf{Compositional}} \\
\underline{\textbf{Add}} & \underline{\textbf{FullAdd}} & \underline{\textbf{Matrix}} & \underline{\textbf{LSTM}} \\
syndicate & syndicate & f(student, representative) & f(worker, representative) \\
representative & f(deputy, representative) & syndicate & f(player, representative) \\
f(worker, representative) & f(student, representative) & f(deputy, representative) & f(crack, dealer) \\
f(deputy, representative) & f(player, representative) & f(worker, representative) & f(company, spokesman) \\
f(student, representative) & f(worker, representative) & f(player, representative) & f(industry, commissioner) \\
\midrule
\rowcolor{lightgray} \multicolumn{4}{c}{\textbf{Paraphrase-based}} \\
\underline{\textbf{Co-occurrence}} & \underline{\textbf{Backtranslation}} \\ 
f(company, representative) & f(worker, representative) \\
f(phone, representative) & f(union, representative) \\
f(union, representative) & f(group, manager) \\
f(marketing, representative) & f(employee, representative) \\
f(labor, representative) & f(student, representative) \\ 
\midrule
\multicolumn{4}{c}{\textit{army officer} (frequent)} \\ \midrule
\rowcolor{lightgray} \multicolumn{4}{c}{\textbf{Distributional}} \\
\multicolumn{4}{c}{army\_captain} \\
\multicolumn{4}{c}{army\_major} \\
\multicolumn{4}{c}{navy\_officer} \\
\multicolumn{4}{c}{army\_general} \\
\multicolumn{4}{c}{army\_lieutenant} \\ 
\midrule
\rowcolor{lightgray} \multicolumn{4}{c}{\textbf{Compositional}} \\
\underline{\textbf{Add}} & \underline{\textbf{FullAdd}} & \underline{\textbf{Matrix}} & \underline{\textbf{LSTM}} \\
army & f(police, commander) & f(police, commander) & f(militia, commander) \\
officer & f(army, troop) & army\_officer & f(police, commander) \\
f(army, battalion) & f(militia, commander) & f(army, troop) & f(opposition, commander) \\
f(army, troop) & f(army, camp) & army\_general & f(military, official) \\
f(army, building) & army\_officer & f(army, camp) & f(comrade, commander) \\ 
\midrule
\rowcolor{lightgray} \multicolumn{4}{c}{\textbf{Paraphrase-based}} \\
\underline{\textbf{Co-occurrence}} & \underline{\textbf{Backtranslation}} \\ 
& f(patrol, officer) & f(army, official) \\
& f(navy, officer) & f(military, spokesman) \\
& f(prison, officer) & f(army, lieutenant) \\
& f(fire, officer) & f(army, chief) \\
& f(police, officer) & f(army, spokesman) \\
\bottomrule
\end{tabular}

    \caption{Top 5 nearest neighbour of two example noun compounds, \textit{syndicate representative} (1 corpus occurrence) and \textit{army officer} (13,924 occurrences) in each composition function. DSM = (word2vec SG, window 5, 300d).}
    \label{tab:example_neighbours}
\end{table*}

\subsection{Constituent Word Embeddings}
\label{sec:rep_constituent_embeddings}

To represent the constituent words, we trained various word embedding algorithms: \textbf{word2vec} \cite{mikolov2013efficient} and \textbf{fastText} \cite{bojanowski2017enriching}, which extends word2vec by adding subword information. We used both the Skip-Gram objective (which predicts the context words given the target word) and the CBOW objective (continuous bag-of-words, predicting the target word from its context).\footnote{We used the Gensim implementation: \url{https://radimrehurek.com/gensim/}} We also trained the \textbf{GloVe} algorithm \cite{pennington2014glove}, which estimates the log-probability of a word pair co-occurrence. All the embeddings were trained on the English Wikipedia dump from January 2018, with various values for the window size (2, 5, 10) and the embedding dimension (100, 200, 300). 

\subsection{Implementation Details}
\label{sec:rep_implementation_details}

We implemented the models using the AllenNLP library \cite{Gardner2017AllenNLP} which is based on the PyTorch framework \cite{paszke2017automatic}. To train the DSMs we used the list of 18,856 \emph{compositional} noun compounds from \newcite{tratz2011semantically}.\footnote{Omitting 351 noun compounds belonging to the \textsc{lexicalized}, \textsc{personal\_name}, and \textsc{personal\_title} classes.} We only used binary noun compounds, i.e. consisting of exactly two constituent nouns, and we split them to 80\% train, 10\% test, and 10\% validation sets.

For the sake of simplicity, for the remainder of the paper we will refer to the training objective and architecture combination as the ``representation'', and a trained instance of the representation, with a choice of underlying word embeddings (algorithm, dimension, and window), as a DSM. 

\section{Experiments}
\label{sec:experiments}
We compare the various representations in 3 experiments: an analysis of the nearest neighbours of each noun compound vector (Section~\ref{sec:exp_nearest_neighbour}), an evaluation on property prediction (Section~\ref{sec:exp_attributes}), and an evaluation on noun compound relation classification (Section~\ref{sec:exp_noun_compound_classification}). 

\subsection{Nearest Neighbour Analysis}
\label{sec:exp_nearest_neighbour}

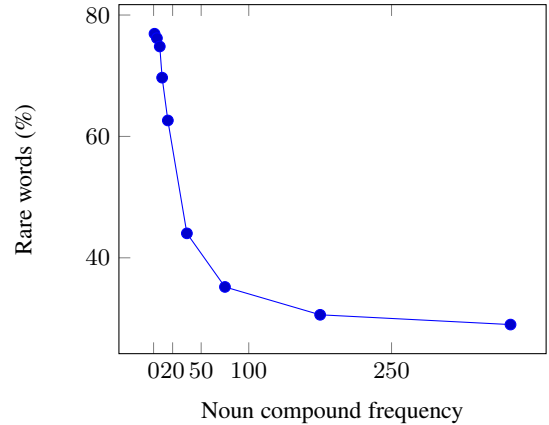
\begin{figure}
    \centering
    \small
    \begin{tikzpicture}
\begin{axis}[
    xtick={0, 20, 50, 100, 250},
    legend style={font=\tiny},
    ytick pos=left,
    % xtick pos=bottom,
    xlabel={Noun compound frequency},
    ylabel={Rare words (\%)},
    width=0.45\textwidth
]
\addplot table [x=freq, y=percent, col sep=comma] {data.csv};
\end{axis}
\end{tikzpicture}
    \caption{Averaged percent of rare words (less than 10 occurrences) among the 10 nearest neighbours of a noun compound with a given corpus frequency.}
    \label{fig:dist_rare_terms}
\end{figure}

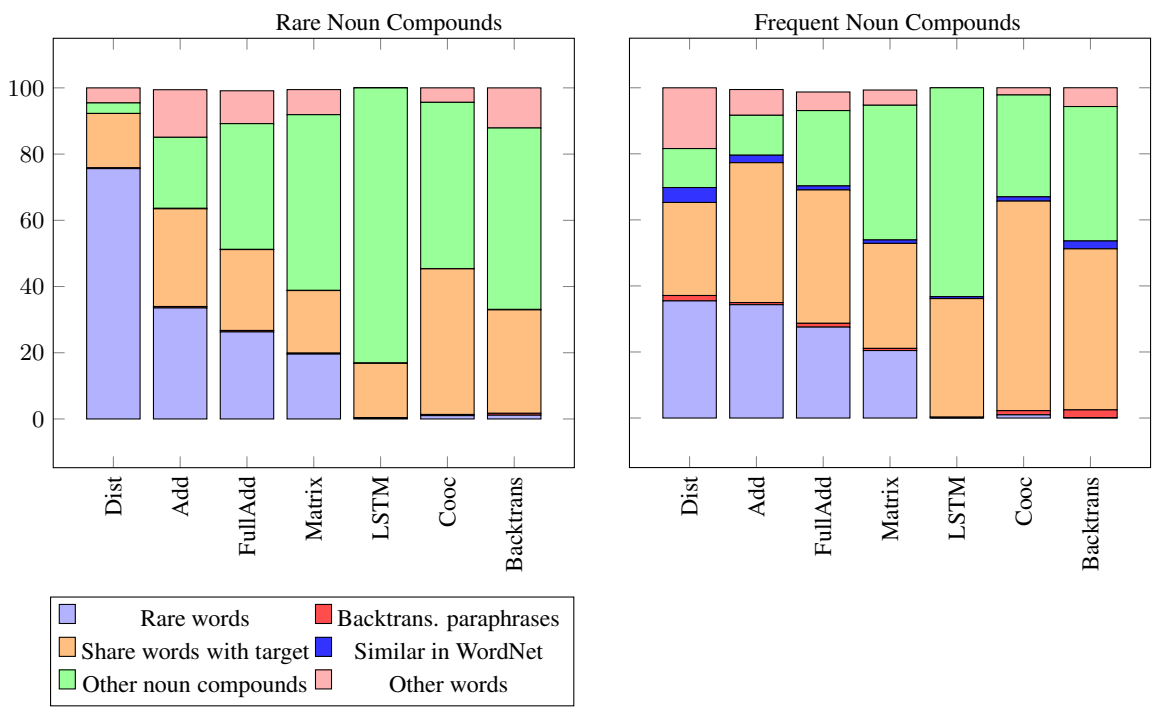
\begin{figure*}[t]
    \centering
    \small
    \begin{tabular}{cc}
     ~~~~~~~~~~~~~~~~~~~~~~~~~~~~~~~~~Rare Noun Compounds & Frequent Noun Compounds  \\
     \begin{tikzpicture}
\begin{axis}[
    ybar stacked,
	bar width=20pt,
	legend style={cells={align=left}},
    enlargelimits=0.15,
    legend style={at={(1,-0.3)}, legend columns=2},
    symbolic x coords={dist, add, fulladd, matrix, lstm, cooc, back},
    xticklabels = {Dist, Add, FullAdd, Matrix, LSTM, Cooc, Backtrans},
    xtick=data,
    x tick label style={rotate=90, anchor=east}
] 
\addplot +[color=black, fill=blue!30] coordinates {(dist,75.61) (add,33.53) 
  (fulladd,26.29) (matrix,19.60) (lstm,0.25) (cooc,1.01) (back,1.14)};
\addplot +[color=black, fill=red!70] coordinates {(dist,0.27) (add,0.44) 
  (fulladd,0.44) (matrix,0.29) (lstm,0.08) (cooc,0.35) (back,0.57)};
\addplot +[color=black, fill=orange!50] coordinates {(dist,16.39) (add,29.50)
  (fulladd,24.42) (matrix,18.89) (lstm,16.47) (cooc,43.98) (back,31.24)};
\addplot +[color=black, fill=blue!80] coordinates {(dist,0.00) (add,0.15) 
  (fulladd,0.07) (matrix,0.06) (lstm,0.08) (cooc,0.10) (back,0.14)};
\addplot +[color=black, fill=green!40] coordinates {(dist,3.18) (add,21.45) 
  (fulladd,37.97) (matrix,53.06) (lstm,83.10) (cooc,50.19) (back,54.79)};
\addplot +[color=black, fill=red!30] coordinates {(dist,4.55) (add,14.37) 
  (fulladd,9.93) (matrix,7.58) (lstm,0.03) (cooc,4.38) (back,12.12)}; 
\legend{\strut Rare words, \strut Backtrans. paraphrases, \strut Share words with target, \strut Similar in WordNet, \strut Other noun compounds, \strut Other words} 
\end{axis}
\end{tikzpicture} & \raisebox{48pt}[0pt][0pt]{%
\begin{tikzpicture}
\begin{axis}[
    ybar stacked,
	bar width=20pt,
    enlargelimits=0.15,
    legend style={at={(-0.1,0.75)}, anchor=east,legend columns=1},
    symbolic x coords={dist, add, fulladd, matrix, lstm, cooc, back},
    xticklabels = {Dist, Add, FullAdd, Matrix, LSTM, Cooc, Backtrans},
    xtick=data,
    yticklabels={},
    x tick label style={rotate=90, anchor=east}
] 
\addplot +[color=black, fill=blue!30] coordinates {(dist,35.46) (add,34.31) (fulladd,27.53) (matrix,20.43) (lstm,0) (cooc,0.96) (back,0.14)};
\addplot +[color=black, fill=red!70] coordinates {(dist,1.64) (add,0.64) (fulladd,1.15) (matrix,0.65) (lstm,0.27) (cooc,1.27) (back,2.32)};
\addplot +[color=black, fill=orange!50] coordinates {(dist,28.17) (add,42.36) (fulladd,40.38) (matrix,31.8) (lstm,35.89) (cooc,63.45) (back,48.76)};
\addplot +[color=black, fill=blue!80] coordinates {(dist,4.52) (add,2.27) (fulladd,1.22) (matrix,1.04) (lstm,0.58) (cooc,1.29) (back,2.39)};
\addplot +[color=black, fill=green!40] coordinates {(dist,11.8) (add,12.09) (fulladd,22.81) (matrix,40.79) (lstm,63.26) (cooc,30.85) (back,40.71)};
\addplot +[color=black, fill=red!30] coordinates {(dist,18.40) (add,7.8) (fulladd,5.63) (matrix,4.61) (lstm,0) (cooc,2.19) (back,5.69)}; 
\end{axis}
\end{tikzpicture}
} \\
\end{tabular}
    \caption{Categories of the top 10 neighbors of each target compound, for the 100 most rare compounds in the test set (first row) and the 100 most frequent compounds in the test set (second row). Best viewed in color.}
    \label{fig:top_k_bar_chart}
\end{figure*}

Similarly to \citeauthor{boleda-et-al:2013:IWCS2013}'s \citeyearpar{boleda-et-al:2013:IWCS2013} analysis for adjective-noun compositions, we compute the 10 nearest neighbors of each noun compound in the test set and analyze the outputs. Table~\ref{tab:example_neighbours} exemplifies the nearest neighbours of two noun compounds in each representation, setting the DSM to (word2vec SG, window 5, 300d). 

\subsubsection{Observed vs. Composed} 

The nearest distributional neighbours of \textit{syndicate representative} in Table~\ref{tab:example_neighbours} demonstrate the well known fact that the distributional embeddings of rare terms are of low quality. The goal of the composition functions is to provide meaningful representations for ad-hoc, possibly rare compositions of nouns. They are learned as an approximation of the observed (distributional) representations of frequent noun compounds. How frequent should a noun compound be for its observed representation to be preferred over the compositional one? For example, the nearest neighbours of \textit{army officer}, a very frequent term, indicate that its distributional embedding is meaningful.\footnote{\newcite{boleda-et-al:2013:IWCS2013} found that in the case of adjective-noun compositions, observed vectors were preferred for frequent compositions, and compositional vectors for rare ones.} 

To get an approximate answer to this question, we compute the percentage of rare words (words which occurred less than 10 times in the corpus) among the 10 nearest neighbours of each noun compound, using the distributional DSMs. We average the percents across the various word embedding algorithms, dimensions, and windows. Figure~\ref{fig:dist_rare_terms} plots the percentage of rare neighbours by noun compound frequency. While the percent of rare words quickly drops from 75\% after only a few occurrences, even noun compounds with more than 250 occurrences have around 30\% of rare neighbours.  

\subsubsection{Neighbour Types} 

We focus on the 100 most frequent compounds (between 3,235 occurrences: \textit{city manager}, and 47,866 occurrences: \textit{ball player}) and the 100 most rare compounds (from one occurrence, e.g. \textit{chief joker}, to 6 occurrences, e.g. \textit{coat shopping}). 

We categorize the neighbours of a target compound into 6 categories, as exemplified for the noun compound \textit{street level}: (1) rare words (\textit{3bf}); (2) other noun compounds which are included in its ``backtranslation'' paraphrases (\textit{ground floor}); (3) the compound's constituents or other noun compounds that share a constituent with it (e.g. \textit{street}, \textit{level}, and \textit{sea level}); (4) words or noun compounds which have high WordNet similarity with the compound\footnote{Specifically, we used the Wu-Palmer similarity \cite{wu1994verbs}, which returns a score denoting how similar two synsets are, based on the depth of their most specific ancestor in the WordNet taxonomy. We took the highest score among all the different synsets of each term, and considered a high score as $>0.25$.}; (5) other noun compounds (\textit{parking garage}); and (6) other words (\textit{stairs}). Figure~\ref{fig:top_k_bar_chart} shows the charts of categories for each representation, averaged across DSMs.

Figure~\ref{fig:top_k_bar_chart} shows that for the compositional representations (Add, FullAdd, Matrix), between 20\% and a third of the neighbours are rare words. The percent of rare words decreases as the composition function has more parameters.\footnote{The percents are similar for frequent and rare noun compounds. This is expected because once the composition function has been learned, the frequency of a test compound has no importance.} The nearest neighbours also typically include trivial neighbours, such as the constituents and other compounds that share a constituent with the target compound (19-30\% for rare compounds and 32-43\% for frequent ones). Overall, at least a half of the neighbours are trivial or meaningless. Most of the other neighbours are other noun compounds which have not been judged for correctness. 

LSTM, Co-occurrence, and Backtranslation all use an LSTM to encode the noun compounds. Although their training objectives are different, they all tend to produce noun compound vectors which are very different from those of single words. This results in nearest neighbour lists which consist of mostly other compounds, either with or without shared constituents. 

Very few neighbours were backtranslation paraphrases: less than 1\% for most representations, and 2.32\% for backtranslation of frequent compounds. 

\begin{table*}
    \centering
    \small
    \begin{tabular}{cccccc}
\toprule
\textbf{Representation} & \textbf{Used for transportation} & \textbf{Is a weapon} & \textbf{Is round} & \textbf{Has various colors} & \textbf{Made of metal} \\ \midrule
Distributional & $48.0 \pm 12.6$  & $\mathbf{57.3 \pm 14.8}$  & $24.8 \pm 8.9$  & $42.0 \pm 12.5$  & $41.3 \pm 12.0$  \\  \midrule
Add & $55.8 \pm 13.5$  & $30.3 \pm 20.1$  & $46.2 \pm 13.2$  & $41.8 \pm 13.1$  & $55.1 \pm 14.1$  \\  
FullAdd & $55.9 \pm 13.4$  & $36.8 \pm 17.3$  & $44.0 \pm 13.0$  & $48.2 \pm 12.7$  & $52.2 \pm 13.0$  \\  
Matrix & $56.5 \pm 13.9$  & $24.0 \pm 19.1$  & $43.8 \pm 13.4$  & $49.5 \pm 13.3$  & $52.0 \pm 12.9$  \\  
LSTM & $48.3 \pm 15.8$  & $0.0 \pm 0.0$  & $21.7 \pm 17.5$  & $37.2 \pm 18.4$  & $42.1 \pm 18.6$  \\  \midrule
Co-occurrence & $\mathbf{64.2 \pm 14.9}$  & $40.5 \pm 30.1$  & $\mathbf{47.0 \pm 13.0}$  & $\mathbf{56.9 \pm 12.8}$  & $\mathbf{57.6 \pm 12.9}$  \\  
Backtranslation & $58.3 \pm 14.1$  & $54.0 \pm 19.5$  & $42.1 \pm 13.5$  & $52.4 \pm 13.5$  & $57.4 \pm 13.1$  \\  
\bottomrule
\end{tabular}

    \caption{Mean and standard deviation of $F_1$ scores across DSMs, for each representation and property combination. The majority baseline $F_1$ score is 0 for all properties, since it always predicts False.}
    \label{tab:property_mean}
\end{table*}

\begin{table*}
    \centering
    \small
    \begin{tabular}{cccccccc}
\toprule
\textbf{Feature} & \textbf{Representation} & \textbf{Embedding} & \textbf{Window} & \textbf{Dimension} & \textbf{Precision} & \textbf{Recall} & \textbf{$\mathbf{F_1}$} \\ \midrule
Used for transportation & Co-occurrence & word2vec SG & 10 & 300 & $74.5$ & $78.8$ & $76.6$ \\ 
Is a weapon & Backtranslation & word2vec CBOW & 2 & 300 & $71.4$ & $88.2$ & $78.9$ \\ 
Is round & Co-occurrence & word2vec CBOW & 10 & 300 & $56.2$ & $87.1$ & $68.4$ \\ 
Has various colors & Co-occurrence & GloVe & 2 & 200 & $70.6$ & $76.6$ & $73.5$ \\ 
Made of metal & Matrix & word2vec SG & 5 & 300 & $78.6$ & $61.1$ & $68.8$ \\ 
\bottomrule
\end{tabular}

    \caption{The performance of the best setting for each property.}
    \label{tab:property_best}
\end{table*}

For frequent compounds, 1-2\% of the neighbours were considered similar to the target compound in WordNet. We note that this category is meaningless for rare noun compounds since most of them are not in WordNet all.\footnote{WordNet only consists of lexicalized noun compounds, e.g. \textit{olive oil} and \textit{ice cream}, which tend to be frequent.}

\subsection{Property Prediction}
\label{sec:exp_attributes}

Do the various representations capture properties of noun compounds? To answer this question, we create a task in which we need to predict for a given noun compound whether it has a certain property or not. For example, is a \textit{cheese wheel} round? 

\subsubsection{Task Definition and Data} 

We use the McRae Feature Norms dataset \cite{mcrae2005semantic}, which provides, for single words describing concrete nouns, the most salient properties that describe them. We follow the binary classification setting of \newcite{rubinstein-etal-2015-well} in which each task is focused on a single property, and negative instances are (a sample of) the concepts that do not appear with the property.

To augment this data with noun compounds, we first filtered the dataset such that it only contains constituents of noun compounds in our vocabulary. We then selected 5 of the most frequent properties (``a weapon'', ``round'', ``made of metal'', ``used for transportation'', and ``comes in different colors''). For each property, we looked for all the noun compounds that consist of a constituent annotated to holding this property, and manually annotated them to whether they also hold this property. For example, since \textit{apple} is round, we manually judged noun compounds such as \textit{apple pie} (also round), and \textit{apple grower} (not round).\footnote{We note that this semi-automatic data collection procedure might miss some salient properties of noun compounds which are not properties of their constituents.} Finally, we manually added some examples from online lists (e.g. the ``round objects'' list in Wikipedia\footnote{\url{https://commons.wikimedia.org/wiki/Category:Round_objects}}).

We split the data to train (90\% of the single words and 20\% of the noun compounds), validation (10\% of the single words and 20\% of the noun compounds), and test (60\% of the noun compounds). The training sets each contains around 500 instances. For each DSM, we train classifiers on the composed vectors of each given concept (a single word or a noun compound). We train multiple classifiers (logistic regression and SVM, with various L2 regularization values) and select the best performing classifier with respect to the validation $F_1$ score. 

\subsubsection{Results and Analysis}

Table~\ref{tab:property_mean} shows the mean and the standard deviation of $F_1$ scores per representation across DSMs, for each of the properties. 

The co-occurrence function stands out in its performance, and the backtranslation function is often second best. There is no clear preference among the compositional functions, except for the LSTM which is consistently worse than the others. The distributional embeddings typically perform among the worst. This is expected both due to the quality of the embeddings of rare noun compounds (Section~\ref{sec:exp_nearest_neighbour}) and since some of the noun compounds in the data are out-of-vocabulary. In contrast, the other representations compute ad hoc vectors for such noun compounds. 

For the sake of completeness, Table~\ref{tab:property_best} displays the best performing DSM for each property. There is a preference to word2vec and to a higher embedding dimension.

Looking at the errors made by the best model we found a common pattern of false positive errors. Most of them stem from multiple positive training instances that share a constituent with the target noun compound, e.g. predicting that \textit{sprint car} is used for transportation, although its primary purpose is racing, that \textit{kidney stone} is a weapon, that \textit{tomato soup} is round, and that \textit{tar ball} comes in multiple colors. We did not find a common pattern among the false negative errors. 

Finally, although it is tempting to draw general conclusions as to the types of properties (e.g. attributive vs. taxonomic) that each representation captures, we refrain from doing so given the small number of properties we tested.

\subsection{Relation Classification}
\label{sec:exp_noun_compound_classification}

\begin{table*}[!t]
    \centering
    \small
    \begin{tabular}{cccccc}
\toprule
\textbf{Representation} & \specialcell{\textbf{Coarse-grained}\\\textbf{Random}} & \specialcell{\textbf{Coarse-grained}\\\textbf{Lexical}} & \specialcell{\textbf{Fine-grained}\\\textbf{Random}} & \specialcell{\textbf{Fine-grained}\\\textbf{Lexical}} \\ \midrule
Distributional & $44.0 \pm 11.5$  & $30.5 \pm 8.5$  & $40.8 \pm 12.5$  & $24.7 \pm 6.5$  \\  \midrule
Add & $51.9 \pm 10.5$  & $34.7 \pm 7.3$  & $51.5 \pm 10.9$  & $30.7 \pm 5.9$  \\  
FullAdd & $\mathbf{54.5 \pm 10.7}$  & $35.7 \pm 8.0$  & $\mathbf{53.5 \pm 11.0}$  & $28.8 \pm 6.8$  \\  
Matrix & $49.1 \pm 11.3$  & $32.6 \pm 8.1$  & $47.3 \pm 12.1$  & $26.7 \pm 7.2$  \\  
LSTM & $54.0 \pm 11.8$  & $\mathbf{37.5 \pm 8.2}$  & $52.1 \pm 11.9$  & $\mathbf{30.9 \pm 6.6}$  \\  \midrule
Co-occurrence & $49.8 \pm 9.7$  & $31.4 \pm 7.1$  & $47.7 \pm 10.6$  & $24.6 \pm 6.0$  \\  
Backtranslation & $47.2 \pm 7.7$  & $33.5 \pm 6.1$  & $44.6 \pm 8.5$  & $26.7 \pm 5.1$  \\ 
\bottomrule
\end{tabular}

    \caption{Mean and standard deviation of $F_1$ scores across word embeddings, windows and dimensions, for each composition function and dataset combination.}
    \label{tab:classification_mean}
\end{table*}

\begin{table*}[!t]
    \centering
    \small
    \begin{tabular}{cccccccc}
\toprule
\textbf{Dataset} & \textbf{Representation} & \textbf{Embedding} & \textbf{Window} & \textbf{Dimension} & \textbf{Precision} & \textbf{Recall} & $\mathbf{F_1}$ \\ \midrule
Coarse-grained Random & LSTM & Fasttext SG & 2 & 300 & $66.5$ & $66.7$ & $66.2$ \\ 
Coarse-grained Lexical & LSTM & Fasttext SG & 2 & 200 & $50.2$ & $49.0$ & $47.5$ \\ 
Fine-grained Random & LSTM & Fasttext SG & 2 & 300 & $64.6$ & $65.3$ & $63.9$ \\ 
Fine-grained Lexical & Matrix & word2vec SG & 2 & 100 & $39.6$ & $39.8$ & $38.1$ \\ 
\bottomrule
\end{tabular}

    \caption{The performance of the best setting for each noun compound relation classification dataset.}
    \label{tab:classification_best}
\end{table*}

Similarly to \newcite{W16-1604}, we also evaluate the various representations on the noun compound classification task. This is a multiclass classification problem to a pre-defined set of semantic relations, e.g. \textit{morning coffee}: \textsc{time} vs. \textit{coffee cup}: \textsc{contained}. 

\subsubsection{Evaluation Setup} 

We evaluate on the \newcite{tratz2011semantically} dataset, which consists of 19,158 instances, labeled in 37 fine-grained relations or 12 coarse-grained relations. We follow the data splits from \newcite{shwartz-waterson:2018:N18-2}, reporting performance on both the random split and the lexical split, in which there are no shared constituents between the train, validation, and test sets. Since we focus on \emph{compositional} noun compounds, we remove the \textsc{lexicalized} relation (which consists of many non-compositional noun compounds). We also remove the \textsc{personal name} and \textsc{personal title} relations which consist of named entities. We train various classifiers on the vectors obtained by each DSM for a given noun compound, choosing the best performing classifier with respect to the validation $F_1$ score. 

It is important to note that the categorization of noun compounds to a fixed inventory of semantic relations that may hold between their constituents is often subjective, making the data noisy. Previous work suggested that many noun compounds fit into more than one relation, and that some relations in the fine-grained version of the data are overlapping \cite{shwartz-waterson:2018:N18-2}. With that said, this data is still a useful proxy for measuring and comparing the quality of representations.

\subsubsection{Results}

Table~\ref{tab:classification_mean} shows the mean and the standard deviation of $F_1$ scores per representation across DSMs, while Table~\ref{tab:classification_best} displays the best DSM for each dataset. 
 
\paragraph{Compositional functions perform better.} The best performing methods are FullAdd and LSTM. Examination of the per-relation $F_1$ scores shows that Add is, for many relations, the best performing composition function. The poor performance of the distributional DSMs may be attributed to the quality of representations for rare noun compounds, although it was also noted by \newcite{shwartz-waterson:2018:N18-2} that even when the target noun compound has a meaningful distributional vector, its most similar neighbor may have been assigned a different label by the annotators, as in \textit{majority party}: \textsc{equative} vs. \textit{minority party}: \textsc{whole+part\_or\_member\_of} (see the discussion in Section~\ref{sec:discussion}). 

In contrast, it is surprising to see that the paraphrase-based DSMs did not perform as well as the compositional ones. We expected their training objective and data to drive the representations towards capturing more explicit information which could aid the classification; for instance, \textit{glass product} has a ``\textit{product made of glass}'' paraphrase in backtranslation and \textit{night meeting} has a ``\textit{meeting held at night}'' paraphrase in co-occurrence. The mediocre performance may be either due to the sparsity of such explicit paraphrases in the data or due to a sub-optimal training objective. We leave further investigation to future work. 

\paragraph{Smaller windows are preferred.} Table~\ref{tab:classification_best} shows a consistent preference to the small window size. DSMs with small windows are known to capture functional, rather than topical similarity between terms, which could to be beneficial for relation classification. For example, \textit{morning workout} in the train set and \textit{night thunderstorm} in the test set are both annotated to \textsc{Time-Of1}. While they are not topically related, they may appear in similar syntactic constructions related to time, e.g. ``before / after / during the \textit{morning workout} / \textit{night thunderstorm}''.  

\paragraph{Some relations are more challenging than others.} The average per-relation $F_1$ scores by representation varies across relations. In the fine-grained version of the dataset, the worse performance was achieved on the \textsc{partial attribute transfer} relation (2.18). In these noun compounds, the modifier ``transfers'' an attribute to the head, as in \textit{bullet train}, which is a fast train (fast ``like a bullet''). Given the figurative nature of this relation, it is not surprising that the various representations struggle in recognizing it. In contrast, the average performance on the \textsc{measure} relation was 71.25, as it is often enough to recognize that the modifier is a measuring unit (e.g. \textit{hour ride}). These observations are in line with previous work \cite{shwartz-waterson:2018:N18-2}.

\paragraph{Comparison to prior work.} The best previously reported $F_1$ scores on these datasets are: coarse-grained random: 77.5, coarse-grained lexical: 47.8, fine-grained random: 73.9, and fine-grained lexical: 42.9 \cite{shwartz-dagan-2018-paraphrase}. They are achieved by richer models and evaluated on the full inventory of semantic relations. Furthermore, the random splits benefit from ``lexical memorization'', i.e. predicting the relation based on the distribution of training instances sharing a single constituent with the target noun compound \cite[e.g., predicting \textsc{topic} for every compound whose head is \textit{guide};][]{W16-1604,shwartz-waterson:2018:N18-2}. This may enhance the performance of models with direct access to the constituent embeddings (e.g. a classifier trained on their vector concatenation). For the sake of comparing between the various representations, we used only the noun compound vectors as input to the classifier.

\section{Discussion}
\label{sec:discussion}
\paragraph{Limitations.} The main limitation of composition functions is that they rely on the assumption of compositionality, which often does not hold. While in this work we focused on compositional noun compounds, the meaning of many noun compounds is not a straightforward combination of the meanings of their constituents. This happens with figurative noun compounds (e.g. \textit{brain drain}, \textit{family tree}), as well as some highly lexicalized ones (e.g., it is not natural to describe \textit{ice cream} using \textit{ice} and \textit{cream}). 

Some representations only operate on binary noun compounds, while the LSTM based representations are capable of producing vectors for variable-length noun compounds. However, we only tested binary noun compounds. It is not certain that the representations we tested would be able to address the complexity of longer noun compounds, which, among other things, also require uncovering the syntactic head-modifier structure. 

Finally, we used a pre-defined list of noun compounds and did not address identification, which should precede both the training and the inference of the representations. While the criteria for selecting what is considered a noun compound can be strictly syntactic, the decision on whether to use (and train) a distributional embedding for a given noun compound may be based on its frequency. 

\paragraph{Contextualized Word Embeddings} are dynamic word embeddings computed for words given their context sentence \cite{peters-EtAl:2018:N18-1,radford2018improving,devlin2018bert}. They have become increasingly popular last year, outperforming static embeddings across NLP tasks. Supposedly, such representations obviate the need to learn dedicated noun compound representations, as the vector of each constituent is computed given the other constituent. 

Recently, \newcite{shwartz2019still} found that while these representations excel at detecting non-compositional noun compounds, they perform much worse at revealing implicit information such as the relationship between the constituents. Moreover, looking into these models' predictions of substitute constituents shows that even when they recognize a constituent is not used in its literal sense (e.g. in non-compositional compounds), the representation of its (often rare) non-literal sense is not always meaningful. Overall, contextualized word embeddings do not completely solve the problem of obtaining meaningful representations for noun compounds, but they do offer a step forward. 

\section{Conclusions}
\label{sec:conclusion}
We trained numerous noun compound representations and compared their quality through a series of tasks and analyses. Our results confirm that distributional representations lose quality as the frequency of the noun compound in the corpus decreases, making dynamic representations imperative. Among such representations, those with more computational power were preferred. There was no single representation that performed best across tasks. The paraphrase-based representations performed better on property identification, while those trained to approximate the distributional representations performed better on relation classification. Two interesting future research directions would be to design a representation with multiple training objectives, and to build it on top of contextualized word representations.

\section*{Acknowledgments}
The author is supported by the Clore Scholars Programme (2017).

\bibliography{references}

\begin{thebibliography}{33}
\expandafter\ifx\csname natexlab\endcsname\relax\def\natexlab#1{#1}\fi

\bibitem[{Bannard and
  Callison-Burch(2005)}]{bannard-callison-burch-2005-paraphrasing}
Colin Bannard and Chris Callison-Burch. 2005.
\newblock \href {https://doi.org/10.3115/1219840.1219914} {Paraphrasing with
  bilingual parallel corpora}.
\newblock In \emph{Proceedings of the 43rd Annual Meeting of the Association
  for Computational Linguistics ({ACL}{'}05)}, pages 597--604, Ann Arbor,
  Michigan. Association for Computational Linguistics.

\bibitem[{Barzilay and McKeown(2001)}]{P01-1008}
Regina Barzilay and R.~Kathleen McKeown. 2001.
\newblock \href {http://aclweb.org/anthology/P01-1008} {Extracting paraphrases
  from a parallel corpus}.
\newblock In \emph{Proceedings of the 39th Annual Meeting of the Association
  for Computational Linguistics}.

\bibitem[{Bojanowski et~al.(2017)Bojanowski, Grave, Joulin, and
  Mikolov}]{bojanowski2017enriching}
Piotr Bojanowski, Edouard Grave, Armand Joulin, and Tomas Mikolov. 2017.
\newblock Enriching word vectors with subword information.
\newblock \emph{Transactions of the Association for Computational Linguistics},
  5:135--146.

\bibitem[{Boleda et~al.(2013)Boleda, Baroni, Pham, and
  McNally}]{boleda-et-al:2013:IWCS2013}
Gemma Boleda, Marco Baroni, The~Nghia Pham, and Louise McNally. 2013.
\newblock \href {http://www.aclweb.org/anthology/W13-0104} {Intensionality was
  only alleged: {O}n {A}djective-noun composition in distributional semantics}.
\newblock In \emph{Proceedings of the 10th International Conference on
  Computational Semantics (IWCS 2013) -- Long Papers}, pages 35--46, Potsdam,
  Germany. Association for Computational Linguistics.

\bibitem[{Brants and Franz(2006)}]{brants2006web}
Thorsten Brants and Alex Franz. 2006.
\newblock Web 1t 5-gram version 1.

\bibitem[{Devlin et~al.(2019)Devlin, Chang, Lee, and
  Toutanova}]{devlin2018bert}
Jacob Devlin, Ming-Wei Chang, Kenton Lee, and Kristina Toutanova. 2019.
\newblock Bert: Pre-training of deep bidirectional transformers for language
  understanding.
\newblock In \emph{Proceedings of the 2019 Conference of the North American
  Chapter of the Association for Computational Linguistics: Human Language
  Technologies, Volume 1 (Long Papers)}, Minneapolis, Minnesota. Association
  for Computational Linguistics.

\bibitem[{Dima(2016)}]{W16-1604}
Corina Dima. 2016.
\newblock \href {https://doi.org/10.18653/v1/W16-1604} {On the compositionality
  and semantic interpretation of english noun compounds}.
\newblock In \emph{Proceedings of the 1st Workshop on Representation Learning
  for NLP}, pages 27--39. Association for Computational Linguistics.

\bibitem[{Dinu et~al.(2013)Dinu, Pham, and Baroni}]{dinu-pham-baroni:2013:CVSC}
Georgiana Dinu, Nghia~The Pham, and Marco Baroni. 2013.
\newblock \href {http://www.aclweb.org/anthology/W13-3206} {General estimation
  and evaluation of compositional distributional semantic models}.
\newblock In \emph{Proceedings of the Workshop on Continuous Vector Space
  Models and their Compositionality}, pages 50--58, Sofia, Bulgaria.
  Association for Computational Linguistics.

\bibitem[{Ganitkevitch et~al.(2013)Ganitkevitch, Van~Durme, and
  Callison-Burch}]{N13-1092}
Juri Ganitkevitch, Benjamin Van~Durme, and Chris Callison-Burch. 2013.
\newblock \href {http://aclweb.org/anthology/N13-1092} {P{P}{D}{B}: The
  paraphrase database}.
\newblock In \emph{Proceedings of the 2013 Conference of the North American
  Chapter of the Association for Computational Linguistics: Human Language
  Technologies}, pages 758--764. Association for Computational Linguistics.

\bibitem[{Gardner et~al.(2018)Gardner, Grus, Neumann, Tafjord, Dasigi, F.~Liu,
  Peters, Schmitz, and Zettlemoyer}]{Gardner2017AllenNLP}
Matt Gardner, Joel Grus, Mark Neumann, Oyvind Tafjord, Pradeep Dasigi, Nelson
  F.~Liu, Matthew Peters, Michael Schmitz, and Luke Zettlemoyer. 2018.
\newblock \href {http://www.aclweb.org/anthology/W18-2501} {Allen{N}{L}{P}: {A}
  deep semantic natural language processing platform}.
\newblock In \emph{Proceedings of Workshop for NLP Open Source Software
  (NLP-OSS)}, pages 1--6, Melbourne, Australia. Association for Computational
  Linguistics.

\bibitem[{Girju(2007)}]{girju:2007:ACLMain}
Roxana Girju. 2007.
\newblock \href {http://www.aclweb.org/anthology/P07-1072} {Improving the
  interpretation of noun phrases with cross-linguistic information}.
\newblock In \emph{Proceedings of the 45th Annual Meeting of the Association of
  Computational Linguistics}, pages 568--575, Prague, Czech Republic.
  Association for Computational Linguistics.

\bibitem[{Hochreiter and Schmidhuber(1997)}]{hochreiter1997long}
Sepp Hochreiter and J{\"u}rgen Schmidhuber. 1997.
\newblock Long short-term memory.
\newblock \emph{Neural computation}, 9(8):1735--1780.

\bibitem[{Kim and Baldwin(2006)}]{kim2006interpreting}
Su~Nam Kim and Timothy Baldwin. 2006.
\newblock Interpreting semantic relations in noun compounds via verb semantics.
\newblock In \emph{Proceedings of the COLING/ACL on Main conference poster
  sessions}, pages 491--498. Association for Computational Linguistics.

\bibitem[{Mallinson et~al.(2017)Mallinson, Sennrich, and
  Lapata}]{mallinson-sennrich-lapata:2017:EACLlong}
Jonathan Mallinson, Rico Sennrich, and Mirella Lapata. 2017.
\newblock Paraphrasing revisited with neural machine translation.
\newblock In \emph{Proceedings of the 15th Conference of the European Chapter
  of the Association for Computational Linguistics: Volume 1, Long Papers},
  pages 881--893, Valencia, Spain. Association for Computational Linguistics.

\bibitem[{McRae et~al.(2005)McRae, Cree, Seidenberg, and
  McNorgan}]{mcrae2005semantic}
Ken McRae, George~S Cree, Mark~S Seidenberg, and Chris McNorgan. 2005.
\newblock Semantic feature production norms for a large set of living and
  nonliving things.
\newblock \emph{Behavior research methods}, 37(4):547--559.

\bibitem[{Mikolov et~al.(2013)Mikolov, Chen, Corrado, and
  Dean}]{mikolov2013efficient}
Tomas Mikolov, Kai Chen, Greg Corrado, and Jeffrey Dean. 2013.
\newblock Efficient estimation of word representations in vector space.
\newblock In \emph{International Conference on Learning Representations
  (ICLR)}.

\bibitem[{Mitchell and Lapata(2010)}]{mitchell2010composition}
Jeff Mitchell and Mirella Lapata. 2010.
\newblock Composition in distributional models of semantics.
\newblock \emph{Cognitive science}, 34(8):1388--1429.

\bibitem[{Paszke et~al.(2017)Paszke, Gross, Chintala, Chanan, Yang, DeVito,
  Lin, Desmaison, Antiga, and Lerer}]{paszke2017automatic}
Adam Paszke, Sam Gross, Soumith Chintala, Gregory Chanan, Edward Yang, Zachary
  DeVito, Zeming Lin, Alban Desmaison, Luca Antiga, and Adam Lerer. 2017.
\newblock Automatic differentiation in {P}y{T}orch.
\newblock In \emph{Autodiff Workshop, NIPS 2017}.

\bibitem[{Pennington et~al.(2014)Pennington, Socher, and
  Manning}]{pennington2014glove}
Jeffrey Pennington, Richard Socher, and Christopher~D. Manning. 2014.
\newblock \href {http://www.aclweb.org/anthology/D14-1162} {Glove: {G}lobal
  {V}ectors for word representation}.
\newblock In \emph{Empirical Methods in Natural Language Processing (EMNLP)},
  pages 1532--1543.

\bibitem[{Peters et~al.(2018)Peters, Neumann, Iyyer, Gardner, Clark, Lee, and
  Zettlemoyer}]{peters-EtAl:2018:N18-1}
Matthew Peters, Mark Neumann, Mohit Iyyer, Matt Gardner, Christopher Clark,
  Kenton Lee, and Luke Zettlemoyer. 2018.
\newblock \href {http://www.aclweb.org/anthology/N18-1202} {Deep contextualized
  word representations}.
\newblock In \emph{Proceedings of the 2018 Conference of the North American
  Chapter of the Association for Computational Linguistics: Human Language
  Technologies, Volume 1 (Long Papers)}, pages 2227--2237, New Orleans,
  Louisiana. Association for Computational Linguistics.

\bibitem[{Poliak et~al.(2017)Poliak, Rastogi, Martin, and
  Van~Durme}]{poliak-EtAl:2017:EACLshort}
Adam Poliak, Pushpendre Rastogi, M.~Patrick Martin, and Benjamin Van~Durme.
  2017.
\newblock \href {http://www.aclweb.org/anthology/E17-2081} {Efficient,
  compositional, order-sensitive {N}-gram embeddings}.
\newblock In \emph{Proceedings of the 15th Conference of the European Chapter
  of the Association for Computational Linguistics: Volume 2, Short Papers},
  pages 503--508, Valencia, Spain. Association for Computational Linguistics.

\bibitem[{Radford et~al.(2018)Radford, Narasimhan, Salimans, and
  Sutskever}]{radford2018improving}
Alec Radford, Karthik Narasimhan, Tim Salimans, and Ilya Sutskever. 2018.
\newblock Improving language understanding by generative pre-training.
\newblock \emph{URL https://s3-us-west-2. amazonaws.
  com/openai-assets/research-covers/language-unsupervised/language\_
  understanding\_paper. pdf}.

\bibitem[{Rubinstein et~al.(2015)Rubinstein, Levi, Schwartz, and
  Rappoport}]{rubinstein-etal-2015-well}
Dana Rubinstein, Effi Levi, Roy Schwartz, and Ari Rappoport. 2015.
\newblock \href {https://doi.org/10.3115/v1/P15-2119} {How well do
  distributional models capture different types of semantic knowledge?}
\newblock In \emph{Proceedings of the 53rd Annual Meeting of the Association
  for Computational Linguistics and the 7th International Joint Conference on
  Natural Language Processing (Volume 2: Short Papers)}, pages 726--730,
  Beijing, China. Association for Computational Linguistics.

\bibitem[{Shwartz and Dagan(2018)}]{shwartz-dagan-2018-paraphrase}
Vered Shwartz and Ido Dagan. 2018.
\newblock \href {https://www.aclweb.org/anthology/P18-1111} {Paraphrase to
  explicate: Revealing implicit noun-compound relations}.
\newblock In \emph{Proceedings of the 56th Annual Meeting of the Association
  for Computational Linguistics (Volume 1: Long Papers)}, pages 1200--1211,
  Melbourne, Australia. Association for Computational Linguistics.

\bibitem[{Shwartz and Dagan(2019)}]{shwartz2019still}
Vered Shwartz and Ido Dagan. 2019.
\newblock Still a pain in the neck: Evaluating text representations on lexical
  composition.
\newblock In \emph{Transactions of the Association for Computational
  Linguistics (TACL)}, page (to appear).

\bibitem[{Shwartz and Waterson(2018)}]{shwartz-waterson:2018:N18-2}
Vered Shwartz and Chris Waterson. 2018.
\newblock \href {http://www.aclweb.org/anthology/N18-2035} {Olive oil is made
  \emph{of} olives, baby oil is made \emph{for} babies: {I}nterpreting noun
  compounds using paraphrases in a neural model}.
\newblock In \emph{Proceedings of the 2018 Conference of the North American
  Chapter of the Association for Computational Linguistics: Human Language
  Technologies, Volume 2 (Short Papers)}, pages 218--224, New Orleans,
  Louisiana. Association for Computational Linguistics.

\bibitem[{Socher et~al.(2012)Socher, Huval, Manning, and Ng}]{D12-1110}
Richard Socher, Brody Huval, D.~Christopher Manning, and Y.~Andrew Ng. 2012.
\newblock \href {http://aclweb.org/anthology/D12-1110} {Semantic
  compositionality through recursive matrix-vector spaces}.
\newblock In \emph{Proceedings of the 2012 Joint Conference on Empirical
  Methods in Natural Language Processing and Computational Natural Language
  Learning}, pages 1201--1211. Association for Computational Linguistics.

\bibitem[{Tratz(2011)}]{tratz2011semantically}
Stephen Tratz. 2011.
\newblock \emph{Semantically-enriched {P}arsing for {N}atural {L}anguage
  {U}nderstanding}.
\newblock University of Southern California.

\bibitem[{Wieting et~al.(2016)Wieting, Bansal, Gimpel, and
  Livescu}]{wieting2015towards}
John Wieting, Mohit Bansal, Kevin Gimpel, and Karen Livescu. 2016.
\newblock Towards universal paraphrastic sentence embeddings.
\newblock In \emph{International Conference on Learning Representations
  (ICLR)}.

\bibitem[{Wieting and Gimpel(2017)}]{wieting-17-millions}
John Wieting and Kevin Gimpel. 2017.
\newblock Pushing the limits of paraphrastic sentence embeddings with millions
  of machine translations.
\newblock In \emph{arXiv preprint arXiv:1711.05732}.

\bibitem[{Wieting et~al.(2017)Wieting, Mallinson, and
  Gimpel}]{wieting-mallinson-gimpel:2017:EMNLP2017}
John Wieting, Jonathan Mallinson, and Kevin Gimpel. 2017.
\newblock \href {https://www.aclweb.org/anthology/D17-1026} {Learning
  paraphrastic sentence embeddings from back-translated bitext}.
\newblock In \emph{Proceedings of the 2017 Conference on Empirical Methods in
  Natural Language Processing}, pages 274--285, Copenhagen, Denmark.
  Association for Computational Linguistics.

\bibitem[{Wu and Palmer(1994)}]{wu1994verbs}
Zhibiao Wu and Martha Palmer. 1994.
\newblock Verbs semantics and lexical selection.
\newblock In \emph{Proceedings of the 32nd annual meeting on Association for
  Computational Linguistics}, pages 133--138. Association for Computational
  Linguistics.

\bibitem[{Zanzotto et~al.(2010)Zanzotto, Korkontzelos, Fallucchi, and
  Manandhar}]{zanzotto2010estimating}
Fabio~Massimo Zanzotto, Ioannis Korkontzelos, Francesca Fallucchi, and Suresh
  Manandhar. 2010.
\newblock Estimating linear models for compositional distributional semantics.
\newblock In \emph{Proceedings of the 23rd International Conference on
  Computational Linguistics}, pages 1263--1271. Association for Computational
  Linguistics.

\end{thebibliography}
\bibliographystyle{acl_natbib}

\newpage
~
\newpage

\appendix
\section{Noun Compound Classification Labels}
The following table displays the semantic relations in the \newcite{tratz2011semantically} dataset. Each coarse-grained relation (highlighted in gray), is followed by the fine-grained relations that it unites. Each fine-grained relation contains an example noun compound (see Section~\ref{sec:exp_noun_compound_classification}).
\begin{table}[!b]
\small
\begin{tabular}{lll}
\toprule 
\rowcolor{lightgray} \multicolumn{2}{l}{\textsc{cause}} \\
~~~~ & experiencer-of-experience & \textit{company strategy} \\
\rowcolor{lightgray} \multicolumn{2}{l}{\textsc{purpose}} \\
~~~~ & purpose & \textit{labor market} \\
~~~~ & create-provide-generate-sell & \textit{aid center} \\
~~~~ & mitigate\&oppose & \textit{fishing quota} \\
~~~~ & perform\&engage\_in & \textit{acquisition fund} \\
~~~~ & organize\&supervise\&authority & \textit{fire commissioner} \\
\rowcolor{lightgray} \multicolumn{2}{l}{\textsc{time}} \\
~~~~ & time-of1 & \textit{fourth-quarter income} \\
~~~~ & time-of2 & \textit{rating period} \\
\rowcolor{lightgray} \multicolumn{2}{l}{\textsc{loc\_part\_whole}} \\
~~~~ & location & \textit{water spider} \\
~~~~ & whole+part\_or\_member\_of & \textit{society member} \\
\rowcolor{lightgray} \multicolumn{2}{l}{\textsc{attribute}} \\
~~~~ & equative & \textit{winter season} \\
~~~~ & adj-like\_noun & \textit{core tradition} \\
~~~~ & partial\_attribute\_transfer & \textit{lemon soda} \\
\rowcolor{lightgray} \multicolumn{2}{l}{\textsc{other}} \\
~~~~ & measure & \textit{percentage change} \\
~~~~ & lexicalized & \textit{action hero} \\
~~~~ & other & \textit{trade conflict} \\
\rowcolor{lightgray} \multicolumn{2}{l}{\textsc{objective}} \\
~~~~ & objective & \textit{biotechnology research} \\
\rowcolor{lightgray} \multicolumn{2}{l}{\textsc{causal}} \\
~~~~ & subject & \textit{government figure} \\
~~~~ & justification & \textit{genocide trial} \\
~~~~ & creator-provider-cause\_of & \textit{refining margin} \\
~~~~ & means & \textit{car bombing} \\
\rowcolor{lightgray} \multicolumn{2}{l}{\textsc{complement}} \\
~~~~ & relational-noun-complement & \textit{police power} \\
~~~~ & whole+attribute\&feature\&quality\_value\_is\_characteristic\_of & \textit{earth tone} \\
\rowcolor{lightgray} \multicolumn{2}{l}{\textsc{containment}} \\
~~~~ & part\&member\_of\_collection\&config\&series & \textit{stock portfolio} \\
~~~~ & contain & \textit{studio lot} \\
~~~~ & variety\&genus\_of & \textit{tuberculosis strain} \\
~~~~ & amount-of & \textit{work load} \\
~~~~ & substance-material-ingredient & \textit{cedar chalet} \\
\rowcolor{lightgray} \multicolumn{2}{l}{\textsc{owner\_emp\_use}} \\
~~~~ & user\_recipient & \textit{subway platform} \\
~~~~ & employer & \textit{government technocrat} \\
~~~~ & owner-user & \textit{government surplus} \\
\rowcolor{lightgray} \multicolumn{2}{l}{\textsc{topical}} \\
~~~~ & personal\_name & \textit{Sarah Boyle} \\
~~~~ & topic\_of\_cognition\&emotion & \textit{security fear} \\
~~~~ & topic\_of\_expert & \textit{cancer expert} \\
~~~~ & obtain\&access\&seek & \textit{finance plan} \\
~~~~ & personal\_title & \textit{Minister Kennedy} \\
~~~~ & topic & \textit{property deal} \\
\bottomrule
\end{tabular}
% \caption{The semantic relations in the \newcite{tratz2011semantically} dataset. Each coarse-grained relation (highlighted in gray), is followed by the fine-grained relations that it unites. Each fine-grained relation contains an example noun compound (see Section~\ref{sec:exp_noun_compound_classification}).}
\end{table}

\end{document}